\documentclass[11pt]{article}
\usepackage{coling2016}
\usepackage{times}
\usepackage{url}
\usepackage{latexsym}
\usepackage{graphicx}
\usepackage{caption}
\usepackage{subcaption}
\usepackage{multirow}

\title{Pre-Translation for Neural Machine Translation}

\author{Jan Niehues, Eunah Cho, Thanh-Le Ha and Alex Waibel\\
Institute for Anthropomatics  \\
Karlsruhe Institute of Technology, Germany \\
{\small \tt firstname.surname@kit.edu}
}

\date{}

\begin{document}
\maketitle
\begin{abstract}

Recently, the development of neural machine translation (NMT) has significantly improved the translation quality of automatic machine translation. While most sentences are more accurate and fluent than translations by statistical machine translation (SMT)-based systems, in some cases, the NMT system produces translations that have a completely different meaning. This is especially the case when rare words occur.

When using statistical machine translation, it has already been shown that significant gains can be achieved by simplifying the input in a preprocessing step. A commonly used example is the pre-reordering approach.

In this work, we used phrase-based machine translation to pre-translate the input into the target language. Then a neural machine translation system generates the final hypothesis using the pre-translation. Thereby, we use either only the output of the phrase-based machine translation (PBMT) system or a combination of the PBMT output and the source sentence. 

We evaluate the technique on the English to German translation task.
Using this approach we are able to outperform the PBMT system as well as the baseline neural MT system by up to 2 BLEU points. We analyzed the influence of the quality of the initial system on the final result.

\end{abstract}

\section{Introduction}
\label{intro}

In the last years, statistical machine translation (SMT) system generated state-of-the-art performance for most language pairs. Recently, systems using neural machine translation (NMT) were able to outperform SMT systems in several evaluations. These models are able to generate more fluent and accurate translation for most of sentences.

Neural machine translation systems provide the output with high fluency. A weakness of NMT systems, however,  is that they sometimes lose the original meaning of the source words during translation. One example from the first conference on machine translation  (WMT16) test set is the segment in Table \ref{example}.

The English word \textit{goalie} is not translated to the correct German word \textit{Torwart}, but to the German word \textit{Gott}, which means \textit{god}. One problem could be that we need to limit the vocabulary size in order to train the model efficiently. We used Byte Pair Encoding (BPE) \cite{sennrich2015} to represent the text using a fixed size vocabulary. In our case the word \textit{goali} is splitted into three parts \textit{go}, \textit{al} and \textit{ie}. Then it is more difficult to transport the meaning to the translation.

In contrast to this, in phrase-based machine translation (PBMT), we do not need to limit the vocabulary and are often able to translate words even if we have seen them only very rarely in the training. In the example mentioned before, for instance, the PBMT system had no problems translating the expression correctly.

On the other hand, official evaluation campaigns \cite{bojar-EtAl:2016:WMT1} have shown that NMT system often create grammatically correct sentence and are able to model the morphologically agreement much better in German. 

The goal of this work is to combine the advantages of neural and phrase-based machine translation systems.  Handling of rare words is an essential aspect to consider when it comes to real-world applications.  The pre-translation framework provides a straightforward way to support such applications. In our approach, we will first translate the input using a PBMT system, which can handle the rare words well. In a second step, we will generate the final translation using an NMT system. This NMT system is able to generate a more fluent and grammatically correct translation. Since the rare words are already handled by the PBMT system, there should be less problems to generate the translation of these words. Using this approach naturally introduces a necessity to handle the potential errors by the PBMT systems.

The remaining of the paper is structured as follows: In the next section we will review the related work. In Section \ref{basics}, we will briefly review the phrase-based and neural approach to machine translation. Section \ref{pre} will introduce the approach presented in this paper to pre-translate the input using a PBMT system. In the following section, we will evaluate the approach and analyze the errors. Finally, we will finish with a conclusion.

\begin{table}
\caption{\label{example} Example translation of NMT}
\centering
\begin{tabular}{ll}
English: & the goalie parried \\
NMT: & der Gott\\
NMT(gloss): & the god \\
\end{tabular}
\end{table}

\section{Related Work}

The idea of linear combining of machine translation systems  using different paradigms has already been used successfully for SMT and rule-based machine translation (RBMT) \cite{dugast-senellart-koehn:2007:WMT,Simard07statisticalphrase-based}. They build an SMT system that is post-editing the output of an RBMT system. Using the combination of SMT and RBMT, they could outperform both single systems.

Those experiments promote the area of automatic post-editing \cite{bojar-EtAl:2015:WMT}. Recently, it was shown that models based on neural MT are very successful in this task \cite{DBLP:journals/corr/Junczys-Dowmunt16}.

For PBMT, there has been several attempts to apply preprocessing in order to improve the performance of the translation system. A commonly used preprocessing step is morphological splitting, like compound splitting in German \cite{ref:koehn2003a}. Another example would be to use pre-reordering in order to achieve more monotone translation \cite{ref:rottmann2007}.

In addition, the usefulness of using the translations of the training data of a PBMT system has been shown. The translations have been used to re-train the translation model \cite{wuebker2010:phraseTraining} or to train additional discriminative translation models \cite{Niehues2013}.

In order to improve the translation of rare words in NMT, authors try to translate words that are not in the vocabulary in a post-processing step \cite{DBLP:conf/acl/LuongSLVZ15}. In \cite{sennrich2015}, a method to split words into sub-word units was presented to limit the vocabulary size. Also the integration of lexical probabilities into NMT was successfully investigated \cite{arthur16}.

\section{Phrase-based and Neural Machine Translation}
\label{basics}

Starting with the initial work on word-based translation system \cite{Brown1993}, phrase-based machine translation \cite{Koehn2003,Och2004} segments the sentence into continuous phrases that are used as basic translation units. This allows for many-to-many alignments.

Based on this segmentation, the probability of the translation is calculated using a log-linear combination of different features:

\begin{equation}
P(e^I,f^I) = \frac{exp(\sum_{n=1}^N \lambda_n h_n(e^I,f^I))}{\sum_{e'^I}exp(\sum_{n=1}^N \lambda_n h_n(e'^I,f^I))}
\end{equation}

In the initial model, the features are based on language and translation model probabilities as well as a few count based features. In advanced PBMT systems, several additional features to better model the translation process have been developed. Especially models using neural networks were able to increase the translation performance.

Recently, state-of-the art performance in machine translation was significantly improved by using neural machine translation. In this approach to machine translation, a recurrent neural network (RNN)-based encoder-decoder architecture is used to transform the source sentence into the target sentence.

In the encoder, an RNN is used to encode the source sentence into a fixed size continuous space representation by inserting the source sentence word-by-word into the network. In a second step, the decoder is initialized by the representation of the source sentence and is then generating the target sequence one word after the other using the last generated word as input for the RNN \cite{SutskeverVL14}.

One main drawback of this approach is that the whole source sentence has to be stored in a fixed-size context vector.  To overcome this problem, \cite{BahdanauCB14} introduced the soft attention mechanism. Instead of only considering the last state of the encoder RNN, they use a weighted sum of all hidden states. Using these weights, the model is able to put attention on different parts of the source sentence depending on the current status of the decoder RNN. In addition, they extended the encoder RNN to a bi-directional one to be able to get information from the whole sentence at every position of the encoder RNN.
A detailed description of the NMT framework can be found in \cite{BahdanauCB14}.

\section{PBMT Pre-translation for NMT (PreMT)}

\begin{figure}
\centering
\caption{Pre-translation methods}
\begin{subfigure}[b]{\textwidth}
	\centering
	\includegraphics[width = 0.7\textwidth]{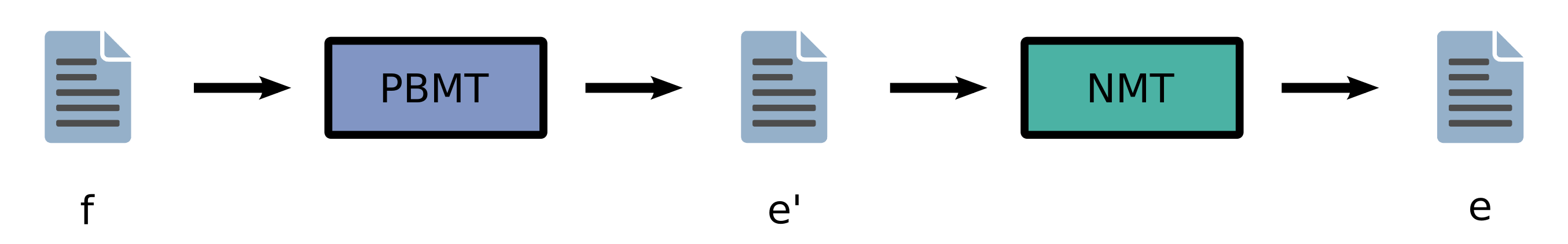}
	\caption{\label{pipeline} Pipeline combination}
\end{subfigure}

\begin{subfigure}[b]{\textwidth}
	\centering
	\includegraphics[width = 0.7\textwidth]{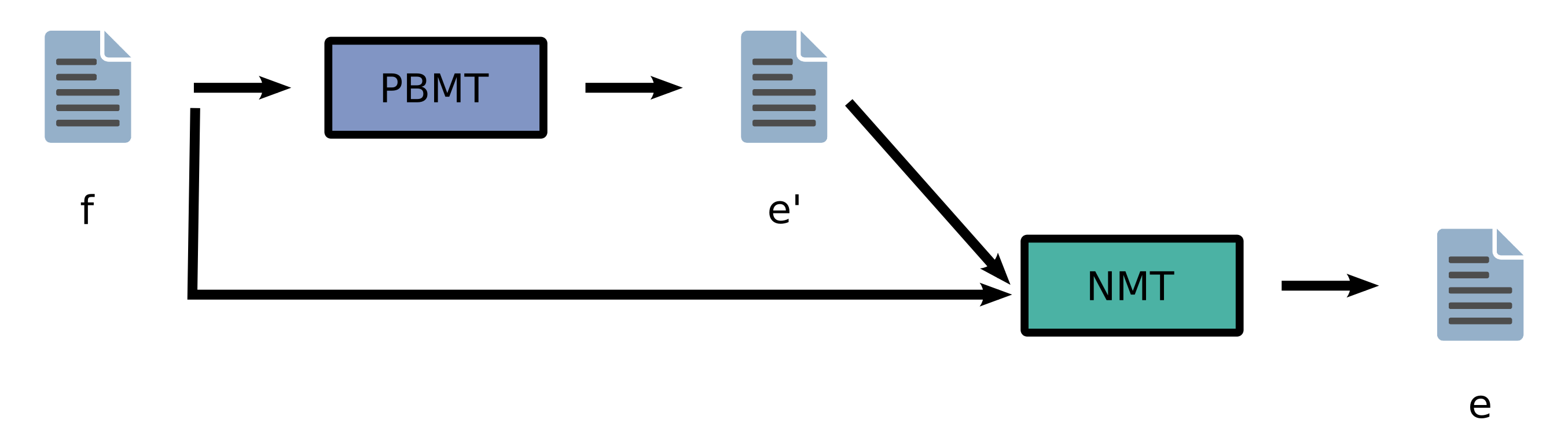}
	\caption{\label{mixed} Mixed Input}
\end{subfigure}

\end{figure}

\label{pre}

In this work, we want to combine the advantages of PBMT and NMT. Using the combined system we should be able to generate a translation for all words that occur at least once in the training data, while maintaining high quality translations for most sentences from NMT. Motivated by several approaches to simplify the translation process for PBMT using preprocessing, we will translate the source as a preprocessing step using the phrase-base machine translation system.

The main translation task is done by the neural machine translation model, which can choose between using the output of the PBMT system or the original input when generate the translation.

\subsection{Pipeline}
\label{pipeline-sec}

In our first attempt, we combined the phrase-based MT and the neural MT in one pipeline as shown in Figure \ref{pipeline}. The input is first processed by the phrase-based machine translation system from the input language $f$ to the target language $e'$.  Since the machine translation system is not perfect, the output of the system may not be correct translation containing errors possibly. Therefore, we will call the output language of the PBMT system $e'$.

In a second step, we will train a neural monolingual translation system, that translates from the output of the PBMT system $e'$ to a better target sentence $e$.

\subsection{Mixed Input}

One drawback of the pipelined approach is that the PBMT system might introduce some errors in the translation that the NMT can not recover from. For example, it is possible that some information from the source sentence gets lost, since the word is entirely deleted during the translation of the PBMT system.

We try to overcome this problem by building an NMT system that does not only take the output of the PBMT system, but also the original source sentence. One advantage of NMT system is that we can easily encode different input information. The architecture of our system is shown in Figure \ref{mixed}.

The implementation of the mixed input for the NMT system is straight forward. Given the source input $f=f_1,\ldots f_I$ and the output of the PBMT system $e'=e'_1,\ldots e'_{J'}$, we generated the input for the NMT system. First, we ensured a non-overlapping vocabulary of $f$ and $e'$ by marking each token in $f$ by a character and $e'$ by different ones. Then both input sequences are concatenated to the input $e^*$ of the NMT system.

Using this representation, the NMT can learn to focus on source word $f_j$ and words $e'_{i'}$ when generating a word $e'_j$.

\subsection{Training}

In both cases, we can no longer train the NMT system on the source language and target language data, but on the output of the PBMT system and the target language data. Therefore, we need to generate translations of the whole parallel training data using the PBMT system.

Due to its ability to use very long phrases, a PBMT system normally performs significantly better on the training data than on unseen test data. This of course will harm the performance of our approach, because the NMT system will underestimate the number of improvements it has to perform on the test data.

In order to limit this effect, we did not use the whole phrase tables when translating the training data. 
If a phrase pair only occurs once,  we cannot learn it from a different sentence pair. Following \cite{Niehues2013}, we removed all phrase pairs that occur only once for the translation of the corpus.

\section{Experiments}

We analyze the approach on the English to German news translation task of the Conference on Statistical Machine Translation (WMT). First, we will describe the system and analyze the translation quality measured in BLEU. Afterwards, we will analyze the performance depending on the frequency of the words and finally show some example translations.

\subsection{System description}

For the pre-translation, we used a PBMT system. In order to analyze the influence of the quality of the PBMT system, we use two different systems, a baseline system and a system with advanced models. The systems were trained on all parallel data available for the WMT 2016\footnote{http://www.statmt.org/wmt16/translation-task.html}. The news commentary corpus, the European parliament proceedings and the common crawl corpus sum up to 3.7M sentences and around 90M words.

In the baseline system, we use three language models, a word-based, a bilingual \cite{ref:niehues2011} and a cluster based language model, using 100 automatically generated clusters using MKCLS \cite{MKCLS}.

The advanced system use pre-reodering  \cite{ref:herrmann2013syntax} and lexicalized reordering. In addition, it uses a discriminative word lexicon \cite{Niehues2013} and a language model trained on the large monolingual data.

Both systems were optimized on the tst2014 using Minimum error rate training \cite{ref:och03mer}. A detailed description of the systems can be found in \cite{ha-EtAl:2016:WMT}.

The neural machine translation was trained using Nematus\footnote{https://github.com/rsennrich/nematus}. For the NMT system as well as for the PreMT system, we used the default configuration. In order to limit the vocabulary size, we use BPE as described in \cite{sennrich2015} with 40K operations. We run the NMT system for 420K iterations and stored a model every 30K iterations. We selected the model that performed best on the development data. For the ensemble system we took the last four models. We did not perform an additional fine-tuning.

The PreMT system was trained on translations of the PBMT system of the corpus and the target side of the corpus. For this translation, we only used the baseline PBMT system.

\subsection{English - German Machine Translation}

The results of all systems are summarized in Table \ref{ENDE-result}. It has to be noted, that the first set, tst2014, has been used as development data for the PBMT system and as validation set for the NMT-based systems.

 \begin{table}[htb]
  \begin{center}
   \begin{tabular}{lccc} \hline \hline
   
    \multirow{2}{*}{System} & Dev/Valid & \multicolumn{2}{c}{Test} \\
    &tst2014&tst2015&tst2016\\
 \hline
 NMT & 20.79 & 23.34 & 27.65\\ 
 NMT Ensemble & 21.42 & 24.03 & 28.89\\ \hline
 PBMT & 19.76 & 21.80 & 26.42 \\
 Advanced PBMT & 21.62 & 23.34 & 28.13 \\  \hline
 Pipeline & 20.56 & 22.04 & 26.75 \\
 Pipeline Advanced & 21.76 & 22.92 & 27.61 \\ \hline 
 Mix & 21.88 & 24.11 & 28.04 \\
 Mix Advanced & 22.53 & 24.37 & 29.62 \\ 
 Mix Advanced Ensemble & 23.16 & 25.35 & 30.67 \\
 \hline

 \hline \hline
   \end{tabular}
 \caption{\label{ENDE-result}Experiments for English$\rightarrow$German}
  \end{center}
 \end{table}

 Using the neural MT system, we reach a BLEU score of 23.34 and 27.65 on tst2015 and tst2016. Using an ensemble system, we can improve the performance to 24.03 and 28.89 respectively. The baseline PBMT system performs 1.5 to 1.2 BLEU points worse than the single NMT system. Using the PBMT system with advanced models, we get the same performance on the tst2015 and 0.5 BLEU points better on tst2016 compared to the NMT system.

First, we build a PreMT system using the pipeline method as described in Section \ref{pipeline-sec}. The system reaches a BLEU score of 22.04 and 26.75 on both test sets. While the PreMT can improve of the baseline PBMT system, the performance is worse than the pure NMT system. So the first approach to combine neural and statistical machine translation is not able the combine the strength of both system. In contrast, the NMT system seems to be not able to recover from the errors done by the SMT-based system.

In a second experiment, we use the advanced PBMT system to generate the translation of the test data.  We did not use it to generate a new training corpus, since the translation is computationally very expensive.  So the PreMT system stays the same, being trained on the translation of the baseline PBMT. However,  it is getting better quality translation in testing. This also leads to an improvement of 0.9 BLEU points on both test sets. Although it is smaller then the difference between the two initial phrase-based translation systems of around 1.5 BLUE points, we are able to improve the translation quality by using a better pre-translation system. It is interesting to see that we can improve the quality of the PreMT system, but improving one component (SMT Pre-Translation), even if we do it only in evaluation and not in training. But the system does not improve over the pure NMT system and even the post editing of the NMT system lowers the performance compared to the initial PBMT system used for pre-translation. 

After evaluating the pipelined system, we performed experiments using the mixed input system. This leads to an improvement in translation quality. Using the baseline PBMT system for per-translation, we perform 0.8 BLEU points better than the purely NMT system on tst2015 and 0.4 BLEU point better on tst2016. It also showed better performance than both PBMT systems on tst2015 and comparable performance with the advanced PBMT on tst2016. So by looking at the original input and the pre-translation, the NMT system is able to recover some of the errors done by the PBMT system and also to prevent errors the NMT does if it is directly translating the source sentence.

Using the advanced PBMT system for input, we can get additional gains of 0.3 and 1.6 BLEU points The system even outperforms the ensemble system on tst2016. The experiments showed that deploying a pre-translation PBMT system with a better quality improves the NMT quality in the mixed input scheme, even when it is used only in testing, not in training.

By using an ensemble of four model, we improve the model by one BLEU point on both test sets, leading to the best results of 25.35 and 30.67 BLEU points. This is 1.3 and 1.8 BLEU points better than the pure NMT ensemble system.

\subsection{System Comparison}

After evaluating the approach, we further analyze the different techniques for machine translation. For this, we compared the single NMT system, the advanced PBMT system and the mixed system using the advanced PBMT system as input.

Out initial idea was that PBMT systems are better for translating rare words, while the NMT is generating more fluent translation. To confirm this assumption, we edited the output of all system. For all analyzed systems, we replaced all target words, which occur in the training data less than $N$ times, by the \textit{UNK} token. For large $N$, we have therefore only the most frequent words in the reference, while for lower $N$ more and more words are used.

\begin{figure}
\centering
\caption{\label{Compare}Compare BLEU score by word frequency}
	\includegraphics[width = 0.7\textwidth]{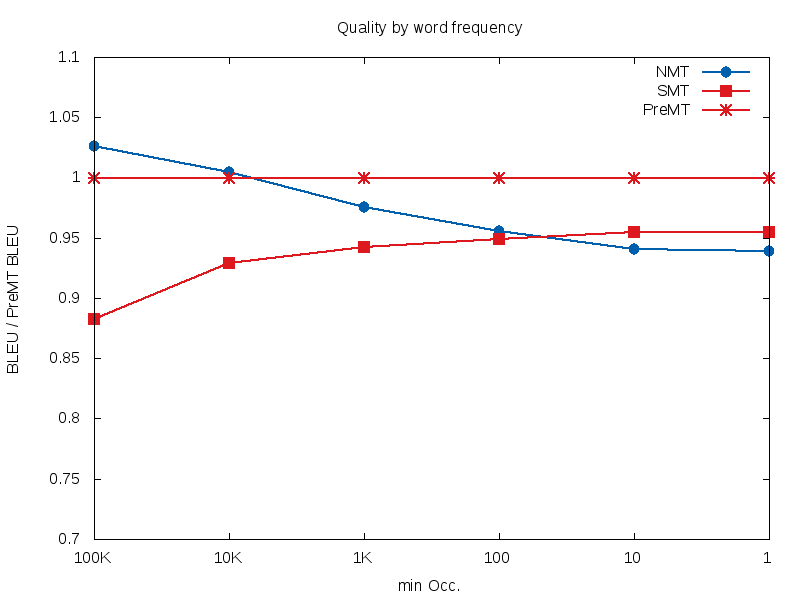}

\end{figure}

The results for $N \in \{1,10,100,1K,10K,100K\}$ are shown in Figure \ref{Compare}. Of course, with lower $N$ we will have fewer \textit{UNK} tokens in the output. Therefore, we normalized the BLEU scores by the performance of the PreMT system.

We can see in the figure, that when $N = 100K$, where only the common words are used, we perform best using the NMT system. The PreMT system performs similar and the PBMT system performs clearly worse. If we now decrease $N$, more and more less frequent words will be considered in the evaluation of the translation quality. Although the absolute BLEU scores raise for all systems, on these less frequent words the PBMT performs better than the NMT system and therefore, finally it even achieves a better performance.

In contrast to this, the PreMT is able to benefit from the pre-translation of the PBMT system and therefore stays better than the PBMT system.

\subsection{Examples}

In Table \ref{example1} we show the output of the PBMT, NMT and PreMT system. First, for the PBMT system, we see a typical error when translating from and to German. The verb of the subclause \textit{parried} is located at the second position in English, but in the German sentence it has to be located at the end of the sentence. The PBMT system is often not able to perform this long-range reordering.

For the NMT system, we see two other errors. Both, the words \textit{goalie} and \textit{parried} are quite rarely in the training data and therefore, they are splitted into several parts by the BPE algorithm. In this case, the NMT makes more errors. For the first word, the NMT system generates a complete wrong translation \textit{Gott (engl. god)} instead of \textit{Torwart}. The second word is just dropped and does not appear in the translation.

The example shows that the pre-translation system prevents both errors. It is generating the correct words \textit{Torwart} and \textit{pariert} and putting them at the correct position in the German sentence.

\begin{table}
\caption{\label{example1} Example sentence translated by different techniques }
\begin{tabular}{ll}
English: & Then with a shot which the \textbf{goalie parried} with his knee in the 35th minute.\\
PBMT: & Dann mit einem Schuss, die der \textbf{Torwart pariert} mit seinem Knie in der 35. Minute.\\
NMT: & Dann mit einem Schuss, den der \textbf{Gott} mit seinem Knie in der 35. Minute. \\
Pre: & Dann mit einem Schuss, das der \textbf{Torwart} mit seinem Knie in der 35. Minute \textbf{pariert}. \\
Pre(gloss): & Then with a shoot, that the goali with his knee in the 35th minute parried. \\
\end{tabular}
\end{table}

\begin{figure}
\centering
\caption{\label{attention}Alignment generated by attention model}
	\includegraphics[width = \textwidth]{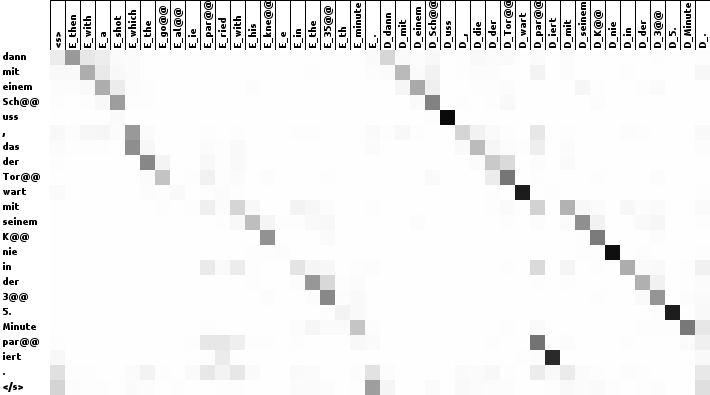}

\end{figure}

To better understand how the pre-translation system is able to generate this translation, we also generated the alignment matrix of the attention model as shown in Figure \ref{attention}. The x-axis shows the input, where the words from the pre-translation are marked by \textit{D\_} and the words from the original source by \textit{E\_}. The y-axis carries the translation. The symbol \textit{@@} marks subword units generated by the BPE algorithm. First, as indicated by the two diagonal lines the model is considering as both inputs, the original source and the pre-translation by the two diagonal lines. 

Secondly, we see that the attention model is mainly focusing on the pre-translation for words that are not common and therefore got splitted into several parts by the BPE, such as \textit{shoot}, \textit{goalie} and \textit{parried}.

A second example, which shows what happens with rare words occur in the source sentence, is shown in Table \ref{example2}. In this case, the word \textit{riot} is not translated but just passed to the target language. This behaviour is helpful for rare words like named entities, but the NMT system is using it also for many words that are not named entities. Other examples for words that were just passed through and not translated are \textit{crossbar} or \textit{vigil}.

\begin{table}
\caption{\label{example2} Example of rare word translation}
\begin{center}
\begin{tabular}{ll}
English: & ... a riot in the stadium.\\
PBMT: & ... einen Aufruhr im Stadion.\\
NMT: & ... einen Riot im Stadion. \\
Pre: & ... einen Aufruhr im Station. \\
Pre (gloss): & ... a riot in\_the stadium.\\
\end{tabular}
\end{center}
\end{table}

\section{Conclusion}

In this paper, we presented a technique to combine phrase-based and neural machine translation. Motivated by success in statistical machine translation, we used phrase-based machine translation to pre-translate the input and then we generate the final translation using neural machine translation.

While a simple serial combination of both models could not generate better translation than the neural machine translation system, we are able to improve over neural machine translation using a mixed input. By simple concatenation of the phrase-based translation and the original source as input for the neural machine translation, we can increase the machine translation quality measured in BLEU. The single pre-translated system could even outperform the ensemble NMT system. For the ensemble system, the PreMT system could outperform the NMT system by up to 1.8 BLEU points.

Using the combined approach, we can generate more fluent translation typical for the NMT system, but also translate rare words. These are often more easily translated by PBMT. Furthermore, we are able to improve the overall system performance by improving the individual components. 

\section*{Acknowledgments}
The project leading to this application has received funding from the European Union's Horizon 2020 research and innovation programme under grant agreement n$^\circ$ 645452.  This work was supported by the Carl-Zeiss-Stiftung.

%

\bibliographystyle{acl}
\bibliography{references}

\begin{thebibliography}{}

\bibitem[\protect\citename{Arthur \bgroup et al.\egroup }2016]{arthur16}
Philip Arthur, Graham Neubig, and Satoshi Nakamura.
\newblock 2016.
\newblock Incorporating discrete translation lexicons into neural machine
  translation.
\newblock In {\em Conference on Empirical Methods in Natural Language
  Processing (EMNLP)}, Austin, Texas, USA, November.

\bibitem[\protect\citename{Bahdanau \bgroup et al.\egroup }2014]{BahdanauCB14}
Dzmitry Bahdanau, Kyunghyun Cho, and Yoshua Bengio.
\newblock 2014.
\newblock Neural machine translation by jointly learning to align and
  translate.
\newblock {\em CoRR}, abs/1409.0473.

\bibitem[\protect\citename{Bojar \bgroup et al.\egroup
  }2015]{bojar-EtAl:2015:WMT}
Ond\v{r}ej Bojar, Rajen Chatterjee, Christian Federmann, Barry Haddow, Matthias
  Huck, Chris Hokamp, Philipp Koehn, Varvara Logacheva, Christof Monz, Matteo
  Negri, Matt Post, Carolina Scarton, Lucia Specia, and Marco Turchi.
\newblock 2015.
\newblock Findings of the 2015 workshop on statistical machine translation.
\newblock In {\em Proceedings of the Tenth Workshop on Statistical Machine
  Translation}, pages 1--46, Lisbon, Portugal, September. Association for
  Computational Linguistics.

\bibitem[\protect\citename{Bojar \bgroup et al.\egroup
  }2016]{bojar-EtAl:2016:WMT1}
Ond\v{r}ej Bojar, Rajen Chatterjee, Christian Federmann, Yvette Graham, Barry
  Haddow, Matthias Huck, Antonio Jimeno~Yepes, Philipp Koehn, Varvara
  Logacheva, Christof Monz, Matteo Negri, Aurelie Neveol, Mariana Neves, Martin
  Popel, Matt Post, Raphael Rubino, Carolina Scarton, Lucia Specia, Marco
  Turchi, Karin Verspoor, and Marcos Zampieri.
\newblock 2016.
\newblock Findings of the 2016 conference on machine translation.
\newblock In {\em Proceedings of the First Conference on Machine Translation},
  pages 131--198, Berlin, Germany, August. Association for Computational
  Linguistics.

\bibitem[\protect\citename{Brown \bgroup et al.\egroup }1993]{Brown1993}
Peter~F. Brown, Vincent J.~Della Pietra, Stephen A.~Della Pietra, and Robert~L.
  Mercer.
\newblock 1993.
\newblock The mathematics of statistical machine translation: Parameter
  estimation.
\newblock {\em Comput. Linguist.}, 19(2):263--311, June.

\bibitem[\protect\citename{Dugast \bgroup et al.\egroup
  }2007]{dugast-senellart-koehn:2007:WMT}
Lo{\"\i}c Dugast, Jean Senellart, and Philipp Koehn.
\newblock 2007.
\newblock Statistical post-editing on systran's rule-based translation system.
\newblock In {\em Proceedings of the Second Workshop on Statistical Machine
  Translation}, pages 220--223, Prague, Czech Republic, June. Association for
  Computational Linguistics.

\bibitem[\protect\citename{Ha \bgroup et al.\egroup }2016]{ha-EtAl:2016:WMT}
Thanh-Le Ha, Eunah Cho, Jan Niehues, Mohammed Mediani, Matthias Sperber,
  Alexandre Allauzen, and Alexander Waibel.
\newblock 2016.
\newblock The karlsruhe institute of technology systems for the news
  translation task in wmt 2016.
\newblock In {\em Proceedings of the First Conference on Machine Translation},
  pages 303--310, Berlin, Germany, August. Association for Computational
  Linguistics.

\bibitem[\protect\citename{Herrmann \bgroup et al.\egroup
  }2013]{ref:herrmann2013syntax}
Teresa Herrmann, Jan Niehues, and Alex Waibel.
\newblock 2013.
\newblock Combining {W}ord {R}eordering {M}ethods on different {L}inguistic
  {A}bstraction {L}evels for {S}tatistical {M}achine {T}ranslation.
\newblock In {\em Proceedings of the Seventh Workshop on Syntax, Semantics and
  Structure in Statistical Translation}, Altanta, Georgia, USA.

\bibitem[\protect\citename{Junczys-Dowmunt and
  Grundkiewicz}2016]{DBLP:journals/corr/Junczys-Dowmunt16}
Marcin Junczys-Dowmunt and Roman Grundkiewicz.
\newblock 2016.
\newblock Log-linear combinations of monolingual and bilingual neural machine
  translation models for automatic post-editing.
\newblock In {\em Proceedings of the First Conference on Machine Translation},
  pages 751--758, Berlin, Germany, August. Association for Computational
  Linguistics.

\bibitem[\protect\citename{Koehn and Knight}2003]{ref:koehn2003a}
Philipp Koehn and Kevin Knight.
\newblock 2003.
\newblock {E}mpirical {M}ethods for {C}ompound {S}plitting.
\newblock In {\em EACL}, Budapest, Hungary.

\bibitem[\protect\citename{Koehn \bgroup et al.\egroup }2003]{Koehn2003}
Philipp Koehn, Franz~Josef Och, and Daniel Marcu.
\newblock 2003.
\newblock Statistical phrase-based translation.
\newblock In {\em Proceedings of the 2003 Conference of the North American
  Chapter of the Association for Computational Linguistics on Human Language
  Technology - Volume 1}, NAACL '03, pages 48--54, Stroudsburg, PA, USA.
  Association for Computational Linguistics.

\bibitem[\protect\citename{Luong \bgroup et al.\egroup
  }2015]{DBLP:conf/acl/LuongSLVZ15}
Thang Luong, Ilya Sutskever, Quoc~V. Le, Oriol Vinyals, and Wojciech Zaremba.
\newblock 2015.
\newblock Addressing the rare word problem in neural machine translation.
\newblock In {\em Proceedings of the 53rd Annual Meeting of the Association for
  Computational Linguistics and the 7th International Joint Conference on
  Natural Language Processing of the Asian Federation of Natural Language
  Processing, {ACL} 2015, July 26-31, 2015, Beijing, China, Volume 1: Long
  Papers}, pages 11--19.

\bibitem[\protect\citename{Niehues and Waibel}2013]{Niehues2013}
Jan Niehues and Alex Waibel.
\newblock 2013.
\newblock {A}n {MT} {E}rror-{D}riven {D}iscriminative {W}ord {L}exicon using
  {S}entence {S}tructure {F}eatures.
\newblock In {\em Proceedings of the Eighth Workshop on Statistical Machine
  Translation}, Sofia, Bulgaria.

\bibitem[\protect\citename{Niehues \bgroup et al.\egroup
  }2011]{ref:niehues2011}
Jan Niehues, Teresa Herrmann, Stephan Vogel, and Alex Waibel.
\newblock 2011.
\newblock {W}ider {C}ontext by {U}sing {B}ilingual {L}anguage {M}odels in
  {M}achine {T}ranslation.
\newblock In {\em Sixth Workshop on Statistical Machine Translation (WMT
  2011)}, Edinburgh, Scotland, United Kingdom.

\bibitem[\protect\citename{Och and Ney}2004]{Och2004}
Franz~Josef Och and Hermann Ney.
\newblock 2004.
\newblock The alignment template approach to statistical machine translation.
\newblock {\em Comput. Linguist.}, 30(4):417--449, December.

\bibitem[\protect\citename{Och}1999]{MKCLS}
Franz~Josef Och.
\newblock 1999.
\newblock An {E}fficient {M}ethod for {D}etermining {B}ilingual {W}ord
  {C}lasses.
\newblock In {\em Proceedings of the Ninth Conference of the European Chapter
  of the Association for Computational Linguistics (EACL 1999)}, Bergen,
  Norway.

\bibitem[\protect\citename{Och}2003]{ref:och03mer}
Franz~Josef Och.
\newblock 2003.
\newblock {M}inimum {E}rror {R}ate {T}raining in {S}tatistical {M}achine
  {T}ranslation.
\newblock In {\em 41st Annual Meeting of the Association for Computational
  Linguistics (ACL)}, Sapporo, Japan.

\bibitem[\protect\citename{Rottmann and Vogel}2007]{ref:rottmann2007}
Kay Rottmann and Stephan Vogel.
\newblock 2007.
\newblock {W}ord {R}eordering in {S}tatistical {M}achine {T}ranslation with a
  {POS}-{B}ased {D}istortion {M}odel.
\newblock In {\em Proceedings of the 11th International Conference on
  Theoretical and Methodological Issues in Machine Translation (TMI 2007)},
  Sk\"{o}vde, Sweden.

\bibitem[\protect\citename{Sennrich \bgroup et al.\egroup }2016]{sennrich2015}
Rico Sennrich, Barry Haddow, and Alexandra Birch.
\newblock 2016.
\newblock Neural machine translation of rare words with subword units.
\newblock In {\em Proceedings of the 54th Annual Meeting of the Association for
  Computational Linguistics, {ACL} 2016, August 7-12, 2016, Berlin, Germany,
  Volume 1: Long Papers}.

\bibitem[\protect\citename{Simard \bgroup et al.\egroup
  }2007]{Simard07statisticalphrase-based}
Michel Simard, Cyril Goutte, and Pierre Isabelle.
\newblock 2007.
\newblock Statistical phrase-based post-editing.
\newblock In {\em In Proceedings of NAACL}.

\bibitem[\protect\citename{Sutskever \bgroup et al.\egroup
  }2014]{SutskeverVL14}
Ilya Sutskever, Oriol Vinyals, and Quoc~V Le.
\newblock 2014.
\newblock Sequence to sequence learning with neural networks.
\newblock In Z.~Ghahramani, M.~Welling, C.~Cortes, N.~D. Lawrence, and K.~Q.
  Weinberger, editors, {\em Advances in Neural Information Processing Systems
  27}, pages 3104--3112. Curran Associates, Inc.

\bibitem[\protect\citename{Wuebker \bgroup et al.\egroup
  }2010]{wuebker2010:phraseTraining}
Joern Wuebker, Arne Mauser, and Hermann Ney.
\newblock 2010.
\newblock Training phrase translation models with leaving-one-out.
\newblock In {\em Annual Meeting of the Assoc. for Computational Linguistics},
  pages 475--484, Uppsala, Sweden, July.

\end{thebibliography}

\end{document}